\documentclass[conference]{IEEEtran}
\IEEEoverridecommandlockouts
\usepackage{cite}
\usepackage{amsmath,amssymb,amsfonts}
\usepackage{algorithmic}
\usepackage{graphicx}
\usepackage{textcomp}
\usepackage{xcolor}
\usepackage{multirow}
\usepackage{hyperref}

\usepackage{array}
\usepackage[caption=false,font=normalsize,labelfont=sf,textfont=sf]{subfig}
\usepackage{subcaption}
\usepackage{stfloats}
\usepackage{float}
\usepackage{url}
\usepackage{verbatim}
\usepackage{adjustbox}
\usepackage{balance}

\def\BibTeX{{\rm B\kern-.05em{\sc i\kern-.025em b}\kern-.08em
    T\kern-.1667em\lower.7ex\hbox{E}\kern-.125emX}}
\begin{document}

\makeatletter
\newcommand{\linebreakand}{%
  \end{@IEEEauthorhalign}
  \hfill\mbox{}\par
  \mbox{}\hfill\begin{@IEEEauthorhalign}
}
\makeatother

\title{UAV-Assisted Real-Time Disaster Detection Using Optimized Transformer Model\\
}

\author{\IEEEauthorblockN{Branislava Jankovic}
\IEEEauthorblockA{\textit{Computer Vision Department} \\
\textit{Mohamed Bin Zayed University of}\\
\textit{Artificial Intelligence}\\
Abu Dhabi, United Arab Emirates \\
branislava.jankovic@mbzuai.ac.ae}
\and
\IEEEauthorblockN{Sabina Jangirova}
\IEEEauthorblockA{\textit{Machine Learning Department} \\
\textit{Mohamed Bin Zayed University of}\\
\textit{Artificial Intelligence}\\
Abu Dhabi, United Arab Emirates \\
sabina.jangirova@mbzuai.ac.ae}
\and
\IEEEauthorblockN{Waseem Ullah}
\IEEEauthorblockA{\textit{Machine Learning Department} \\
\textit{Mohamed Bin Zayed University of}\\
\textit{Artificial Intelligence}\\
Abu Dhabi, United Arab Emirates \\
waseemullah@ieee.org}
\linebreakand
\IEEEauthorblockN{Latif U. Khan}
\IEEEauthorblockA{\textit{Machine Learning Department} \\
\textit{Mohamed Bin Zayed University of}\\
\textit{Artificial Intelligence}\\
Abu Dhabi, United Arab Emirates \\
latif.u.khan2@gmail.com,}
\and
\IEEEauthorblockN{Mohsen Guizani}
\IEEEauthorblockA{\textit{Machine Learning Department} \\
\textit{Mohamed Bin Zayed University of}\\
\textit{Artificial Intelligence}\\
Abu Dhabi, United Arab Emirates \\
mguizani@ieee.org}
}

\maketitle

\begin{abstract}
Dangerous surroundings and difficult-to-reach landscapes introduce significant complications for adequate disaster management and recuperation. These problems can be solved by engaging unmanned aerial vehicles (UAVs) provided with embedded platforms and optical sensors. In this work, we focus on enabling onboard aerial image processing to ensure proper and real-time disaster detection. Such a setting usually causes challenges due to the limited hardware resources of UAVs. However, privacy, connectivity, and latency issues can be avoided. We suggest a UAV-assisted edge framework for disaster detection, leveraging our proposed model optimized for onboard real-time aerial image classification. The optimization of the model is achieved using post-training quantization techniques. To address the limited number of disaster cases in existing benchmark datasets and therefore ensure real-world adoption of our model, we construct a novel dataset, DisasterEye, featuring disaster scenes captured by UAVs and individuals on-site. Experimental results reveal the efficacy of our model, reaching high accuracy with lowered inference latency and memory use on both traditional machines and resource-limited devices. This shows that the scalability and adaptability of our method make it a powerful solution for real-time disaster management on resource-constrained UAV platforms.

The code and DisasterEye dataset are available at: \textcolor{red}{https://github.com/Branislava98/TensorRT}.  
\end{abstract}


\section{Introduction}
\IEEEPARstart{O}{ver} many decades, humanity and forests have faced various disasters that have claimed countless lives and inflicted significant financial losses. These events affect both economically developed and underdeveloped regions, underscoring the persistent limitations of current technological solutions aimed at identifying the early onset of disasters as well as remediating their consequences. 

In 2024, the United Arab Emirates registered record-breaking rainfall, reaching 254.8 mm in less than 24 hours \cite{nasa}, resulting in five deaths, the cancellation of 1,244 flights, the suspension of routes between emirates, and 544.6 million dollars of flood damage. At the same time, Balkan countries were affected by a massive increase in forest fires. According to data from the European Forest Fire Information Service (EFFIS) \cite{effis}, Greece had 56 fires in 2023, burning 174,773 hectares of land, causing damage of more than 1.8 billion euros and 3,058,528 tons of CO2 emissions. This emphasizes that many countries must improve disaster management.
 
Disaster management involves a variety of actions designed to minimize the impact of disasters on society and the environment. Early detection of disaster nodes can help reduce their harmful causes and eliminate them more efficiently. Moreover, an adequate ongoing response and effective remediation of the consequences can prevent the disruption of the population's normal rhythm. Disaster management is highly dependent on human intervention, which can be challenging in unpredictable situations and difficult terrains. This limitation can be overcome by adopting UAVs and UAV-based models for the detection, observation, and study of passive and active threats and risks in incident scenes, which could significantly reduce the need for human intervention and human error \cite{Ijaz2023}.

UAVs have the ability to access locations that are difficult or dangerous for humans to reach. When equipped with camera sensors, they enable real-time capture and transfer of high-resolution images. Traditionally, these images are processed using cloud computing frameworks; however, this approach faces significant challenges, including high latency, limited throughput, increased power consumption, and dependency on reliable cloud services, which may not always be available in disaster-prone or remote areas. Therefore, onboard processing is required at the edge. However, UAVs come with limited computational resources and low-power constraints, which can be challenging in performing computer vision tasks due to massive neural network models \cite{Kyrkou}.  In particular, running the network with a large number of parameters and computational overhead on such devices can cause memory shortages and delayed decision-making due to increased latency. This is why existing solutions for disaster management on UAVs hardly rely on shallow networks or compressed CNN models \cite{Kyrkou}, \cite{mogaka2024tinyemergencynet}, \cite{yar2023fire_detection}, \cite{yar2024efficient}, \cite{Ijaz2023}. 

The disadvantage of shallow networks is their insufficient complexity, which limits the learning of valuable features for classification. On the other hand, CNNs struggle with larger receptive fields. Consequently, more complex architectures, such as transformers, need to be considered. This work addresses the problem of onboard disaster classification for UAVs using complex neural networks. We present a UAV-assisted edge framework that trains a transformer-based neural network on different aerial image databases, which is further optimized to provide the best exchange between accuracy and performance and operate on edge devices. Moreover, we tackle the lack of real-world relevance in existing datasets, which mainly focus on certain disaster cases or contain only a few disaster cases, making them unsuitable for real-world adaptation. We construct a novel database with seven distinguishing disaster cases that feature disaster scenes captured by UAVs and individuals on-site.

These are the summarized main contributions of this work:
\begin{enumerate}
    \item{We introduce a UAV-assisted edge framework for disaster detection, leveraging an optimized transformer-based architecture for accurate aerial image classification. Our framework is tested on existing aerial imagery datasets. Additionally, we explore various compression and quantization techniques, providing an in-depth analysis of their impacts on performance.}
    \item{To enhance relevance and realism, we construct a novel dataset DisasterEye, containing real-world scenes captured by UAVs and individuals on-site.}
    \item{Our Proposed model achieved real-time performance, 3$x$ to 5$x$ times faster than the original, with almost similar accuracy. Moreover, we validate results on a resource-constrained device (Jetson Nano) to determine the potential for deploying our solution on UAVs.}
\end{enumerate}

The rest of this paper is structured as follows: Section \ref{brw} provides related works and background. Section \ref{met} explains the methodology, in particular, the training and optimization of the model. In Section \ref{res}, we analyze datasets and results. Finally, Section \ref{con} provides concluding remarks and discusses directions for future work.

\section{Background and Related work}
\label{brw}
This section discusses previous work on disaster classification with UAVs and highlights the background knowledge of Vision Transformers and Optimization techniques. 

One of the first solutions, \cite{Ko2009}, proposed a traditional vision sensor-based fire detection method for an early warning fire monitoring system with an SVM classifier. In the following decade, works such as \cite{Kim2016, Zhao2018} used a neural network based on VGG16 for fire classification, while \cite{Sharma2017, Xiang2017} demonstrated the effectiveness of residual-based neural networks. Shallow architectures specially tailored for fire detection were introduced in \cite{Jadon2019, yar2022optimized, khan2023enhancing,yar2023fire_detection, yar2024efficient, dataset, Kyrkou, mogaka2024tinyemergencynet, Li2020}, and they were focusing on the classification of disaster aerial images. As the field was growing, new datasets were introduced, as well as new problems. The authors in \cite{Rahnemoonfar2021} presented a new high-resolution aerial imagery dataset for post-flood scene understanding, FloodNet. The dataset was further used in \cite{Khose2021} for semi-supervised classification and supervised semantic segmentation. In work \cite{PresaReyes2022}, researchers tackled the problem of low-altitude imagery and used semi-supervised training techniques to learn from noisy, constrained, and erroneous labels, while works presented in \cite{Shoukry2021,Ijaz2023} studied the issue of model size.

From the previous work, we can observe that most of them were based on lightweight convolutional neural networks. CNNs may be easily exported to edge devices but are also limited by the number of learning parameters, making them unable to capture valuable features \cite{Sze2017}.

\begin{figure*}[htbp] 
    \centering
    \includegraphics[width=18cm, height=8cm]{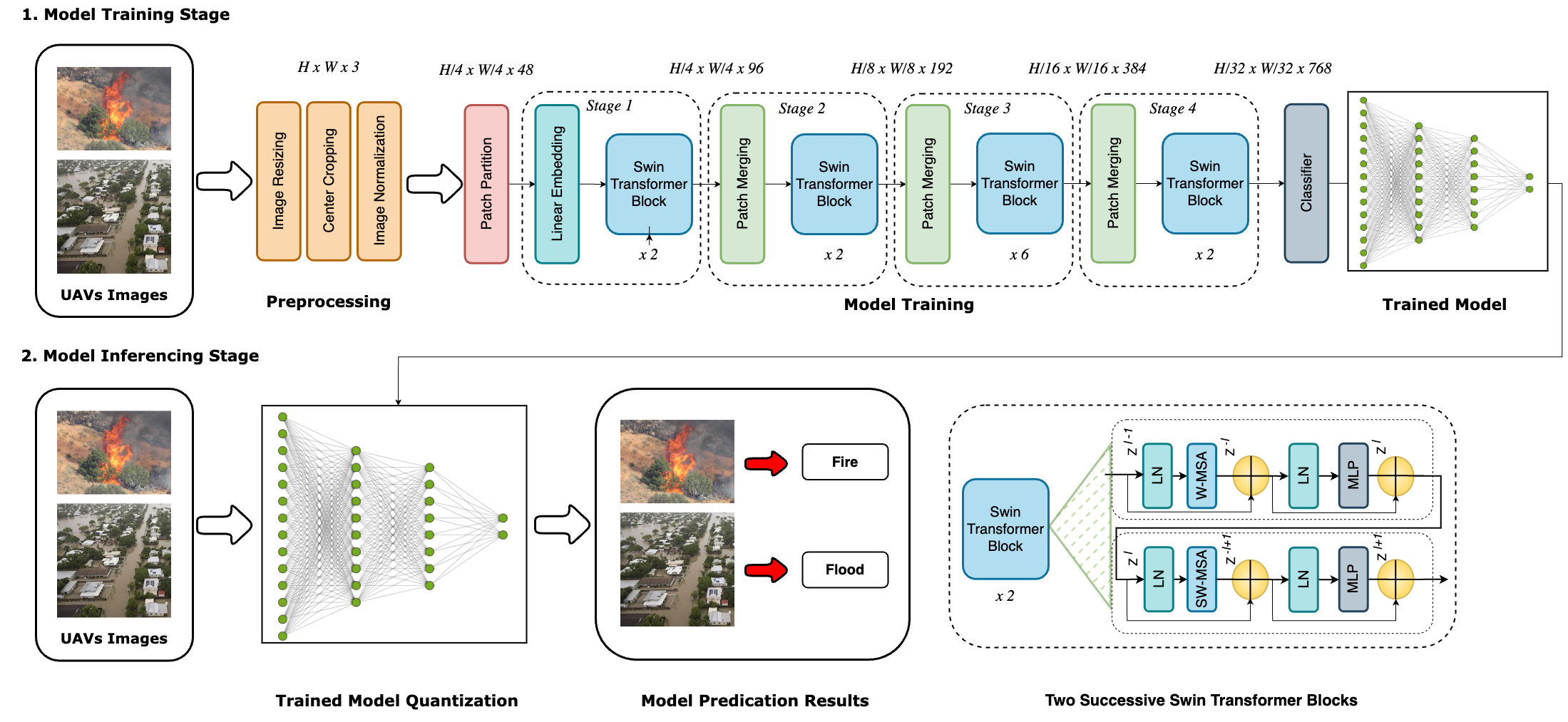} 
    \caption{Overview of the proposed framework: 1) Model training, where UAV-captured images are processed through preprocessing, feature extraction using a backbone transformer model, and training; and 2) Model inference, enabling real-time disaster detection.}
    \label{fig:pipeline}
\end{figure*}
\subsection{\textit{Vision Transformers}}

Based on attention mechanisms, Transformers were initially designed for sequence modeling and transduction tasks \cite{vaswani2017attention}. The ability to effectively capture long-range dependencies guided researchers to utilize them for computer vision tasks. Vision Transformer (ViTs) \cite{dosovitskiy2020image} is the first transformer-based architecture that was applied to the computer vision domain. Thanks to the use of attention mechanisms to capture global receptive fields, they quickly showed effectiveness in computer vision tasks, such as classification \cite{dosovitskiy2020image}, \cite{touvron2021training}, object detection \cite{carion2020end}, \cite{liu2021swin}, semantic segmentation \cite{zheng2021rethinking}, and many more. Nevertheless, the significant number of parameters and the computational overhead of transformers present a challenge during the deployment of resource-constrained hardware devices \cite{lin2023fqvit}. Thus, compression approaches for ViTs are necessary for practical deployments \cite{li2023ivit}.

\subsection{\textit{Optimization Techniques}}

Most existing Post-Training Quantization approaches have been developed and tested on CNNs. However, they lack appropriate handling of transformer-based models with nonlinear arithmetic, such as Softmax, GELU, and LayerNorm. FasterTransformer \cite{nvidia2022fastertransformer} is designed to deliver high-performance inference, but leaves non-linear operations as dequantized floating-point arithmetic. Q-ViT \cite{li2022qvit} takes the widths and scales of the quantization as able-to-learn parameters. Other solutions are ranking loss \cite{liu2021post}, presented to maintain the correct relative order of the quantized attention map, PTQ4ViT \cite{yuan2021ptq4vit} with twin uniform quantization and a Hessian metric for the evaluation of different scaling factors, and FQ-ViT \cite{lin2023fqvit} which introduces two different quantizations specially tailored for LayerNorm and Softmax, but ignores the GELU operation since it is based on I-BERT \cite{kim2021ibert}. Finally, RepQ-ViT \cite{li2022repqvit} addresses the extreme distributions of LayerNorm and Softmax activations, while PSAQ-ViT \cite{li2022psaqvitv2}, \cite{li2022patch} pushes ViT quantization to data-free scenarios based on patch similarity.

\section{Methodology}
\label{met}
\subsection{Problem Formulation}

Although traditional CNNs have been widely used in UAV disaster classification tasks due to their effectiveness on edge devices, they rely on localized receptive fields, which can affect the capture of long-range dependencies essential for complex scene understanding. In contrast, Vision Transformers surpassed this limitation by introducing attention mechanisms \cite{dosovitskiy2020image}. Nevertheless, deployment on edge devices becomes a challenge due to their vast number of parameters and higher computational costs. Our proposed framework builds on these insights by optimizing a transformer-based model specifically for UAV disaster classification tasks, employing PTQ methods to meet the field's real-time, low-power demands of UAV-assisted edge frameworks.

The main steps to develop the disaster classification pipeline are shown in Fig. \ref{fig:pipeline} and divided into two stages: the \textbf{Model Training Stage} that includes preprocessing and training of the model, and the \textbf{Model Inferencing Stage} that incorporates network optimization and inference evaluation.


\subsection{Model Architecture}
We utilize the Swin transformer \cite{liu2021swin} for disaster classification. Its hierarchical architecture with shifting windows allows performance at various scales with linear computational complexity. This enables the model to focus on detailed and global features, which is impossible with CNNs. In the beginning, a patch-splitting module splits RGB images into non-overlapping patches. Each patch represents a "token" that will be further projected into an arbitrary dimension with a linear embedding layer. To produce a hierarchical architecture, patch merging layers reduce the number of "tokens" as the network gets deeper. At the same time, the number of channels increases. Swin Transformer blocks follow the linear embedding layer in the first stage and patch merging layers in the other stages. Their role is to transform features while maintaining the same resolution. Fig. \ref{fig:pipeline} illustrates two successive Swin transformer blocks. In the second block, the standard multi-head self-attention (W-MSA)
module is replaced by a module based on shifted windows, while the rest of the layers remain the same. This was done due to a lack of connections between neighboring non-overlapping windows.


\subsection{Model Optimization}
In this work, we use Post-Training Quantization to optimize our models. PTQ is a technique for lowering computational and memory expenses of running inference. This is done by defining the weights, activations, and attention with low-precision data types such as 16-bit floating point and integer instead of the original 32-bit floating point. The model can achieve faster performance thanks to simpler matrix multiplications by reducing the number of bits. Moreover, it requires less storage memory and consumes less energy. In this work, we are employing the TensorRT quantization technique and compare it with MinMax, EMA \cite{Jacob}, Percentile \cite{Li}, OMSE \cite{Choukroun}, and FQ-Vit \cite{lin2023fqvit}, quantization methods. TensorRT is an SDK developed by NVIDIA specifically for faster performance on NVIDIA GPUs \cite{nvidia2024developer}, using five different optimizations: precision calibration, layer and tensor fusion, kernel auto-tuning, multi-stream executions, and dynamic tensor memory. 

\subsection{Model Inference}

We comprehensively evaluate the inference of the optimized model across two distinct platforms, one of which is resource-constrained. 
Each platform provides unique hardware capabilities that impact inference speed, memory consumption, and overall efficiency, allowing us to assess the adaptability and scalability of our model across different environments.

\section{Experimental Results}
\label{res}

\subsection{Datasets}

Our work utilizes three different datasets: DFAN \cite{yar2022optimized}, AIDER \cite{dataset}, and DisasterEye, our custom dataset. Fig. \ref{fig:images_datasets} shows example images in these datasets.

\begin{figure}[htbpt] 
    \centering
    \includegraphics[width=0.49\textwidth]{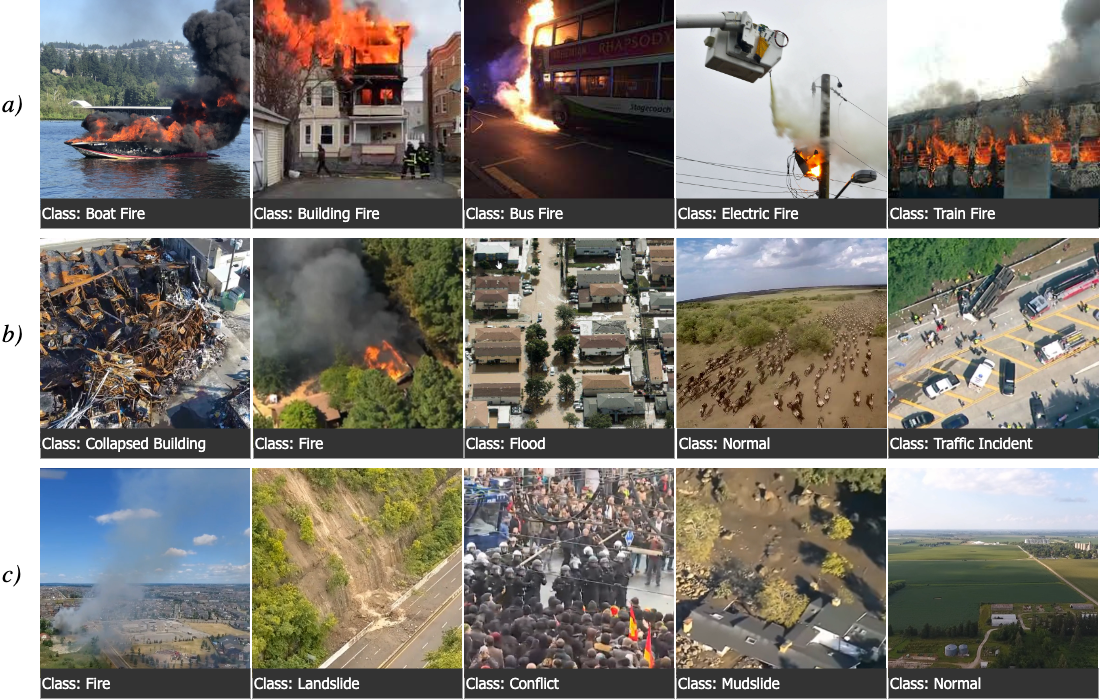} 
    \caption{Samples images of various UAVs based datasets: a) DFAN; b) AIDER; c) DisasterEye.}
    \label{fig:images_datasets}
\end{figure}

The DFAN dataset is a medium-scale database of 3,803 images representing different fire scenarios, divided into 12 imbalanced classes. The photos were sourced from multiple locations, including videos, leading to duplicate images. Due to noisiness and huge class diversity, this dataset hinders the training of models. The AIDER dataset \cite{dataset} is a unique and comprehensive resource for disaster classification tasks. It comprises 6923 images divided into five classes, including four disaster classes: collapsed buildings, fire, flood, and traffic accidents, followed by a normal class. Aerial disaster images were manually collected from various online sources. Each disaster class contains approximately 500 photographs, while the average class has a significantly more extensive set of 4,390 images, which can challenge models on inference.

Previous datasets have only a few disaster cases, or they are focused on only one disaster case, such as a fire. Therefore, the models trained on these databases are unsuitable for real-world applications. Consequently, a benchmark dataset that will resemble real-life scenarios is needed. DisasterEye is our custom dataset that contains images taken with UAVs during or after disasters, as well as images taken from individuals on sight. This dataset consists of 2751 images separated into eight classes: flood, fire, traffic accident, post-earthquake, mudslide, landslide, normal, and conflict. The samples are collected from various sources such as Google Images and YouTube. Similarly to the DFAN dataset in \cite{yar2022optimized}, we uniformly divided the dataset into three sets: 70\% of the data for training, 20\% for validation and the remaining 10\% for testing purposes. Fig. \ref{fig:stat} shows the distribution of instances per class in the training, validation and testing subsets.

\begin{figure}[htbpt] 
    \centering
    \includegraphics[width=0.49\textwidth]{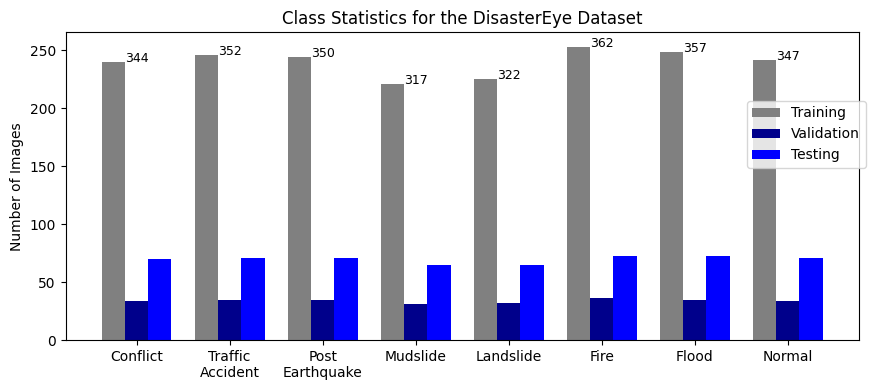} 
    \caption{Class statistics for the DisasterEye dataset.}
    \label{fig:stat}
\end{figure}



\subsection{Evaluation Metrics}

We used the same evaluation metrics: accuracy, precision, recall and f1-score for a fair assessment of the proposed model, evaluating its performance in disaster detection on UAVs and under various quantization strategies. The accuracy expresses the number of correctly classified instances in one dataset, while loss is used to analyze the performance of trained models. The f1-score provides a balance between precision-positive predictive value and recall-model sensitivity. In addition to the metrics mentioned above, we focus on latency, throughput, and model size to evaluate quantization strategies. Latency presents the model's time to complete a single inference. Throughput tells us how many instances can be processed in a specific time frame, in particular, we measure how many frames can be processed in one second (FPS). Finally, the size of the model is essential for deploying the model on edge devices.

\subsection{Training Performance}

Before training, the images of each dataset are divided into training, validation, and testing subsets. Splitting the data is followed by image prepossessing, including resizing, cropping, and normalization to ensure consistency in size, format, and intensity distribution. We fine-tune the transformer-based network trained on ImageNet with our datasets to avoid overfitting and achieve good generalization. All models are fine-tuned until convergence with the Adam optimizer and a learning rate of $1e-5$.
The training is conducted on a desktop computer with an NVIDIA GeForce RTX 2080 12 GB, an Intel Xeon Silver 4215 CPU running at 2.50 GHz, and 12 GB of RAM. For evaluation, we use an NVIDIA T4 GPU on Google Colab, as it offers a more convenient setup for TensorRT installation. We evaluate our proposed model on three different datasets: AIDER, DFAN, and DisasterEye, using categorical cross-entropy loss. Our model performs satisfactorily on the AIDER dataset, while a larger number of classes and a limited number of instances in DFAN and DisasterEye hinder the model's generalization, as shown in Table \ref{tab:opt}.




\subsection{Optimization Performance}
We utilize TensorTR for optimizing our model. TensorRT allows two types of precision calibration: FP16 and INT8. 

Precision calibration with 16-bit reduces the memory footprint, which minimizes the model size roughly by half and gives four times faster inference. However, this solution can affect the models' accuracy. This can be seen from the performance of the quantized model on the DFAN dataset (Table \ref{tab:opt}). The accuracy of the model decreased by approximately 1.9\%. The remaining two datasets do not face such a drop in accuracy. 

Plain INT8 quantization in TensorRT with default dynamic ranges is used to optimize our model to INT8 precision. We do not observe a notable drop in accuracy (more than 0.3\%) compared to FP16 quantization on our datasets. However, there is a significant drop in the model size, which is now reduced by roughly 70\%. At the same time, latency is slightly increased compared to TensorRT FP16 optimization due to the need for precision scaling and zero-point adjustments. Overall, both FP16 and INT8 quantization with TensorRT achieve real-time performance which can be seen from Table \ref{tab:opt}.

\subsection{Visual Results Analysis}
Fig. \ref{fig:inference} shows the visual results of the performance of our quantized model on our datasets. 
\begin{figure}[h] 
    \centering
    \includegraphics[width=0.49\textwidth]{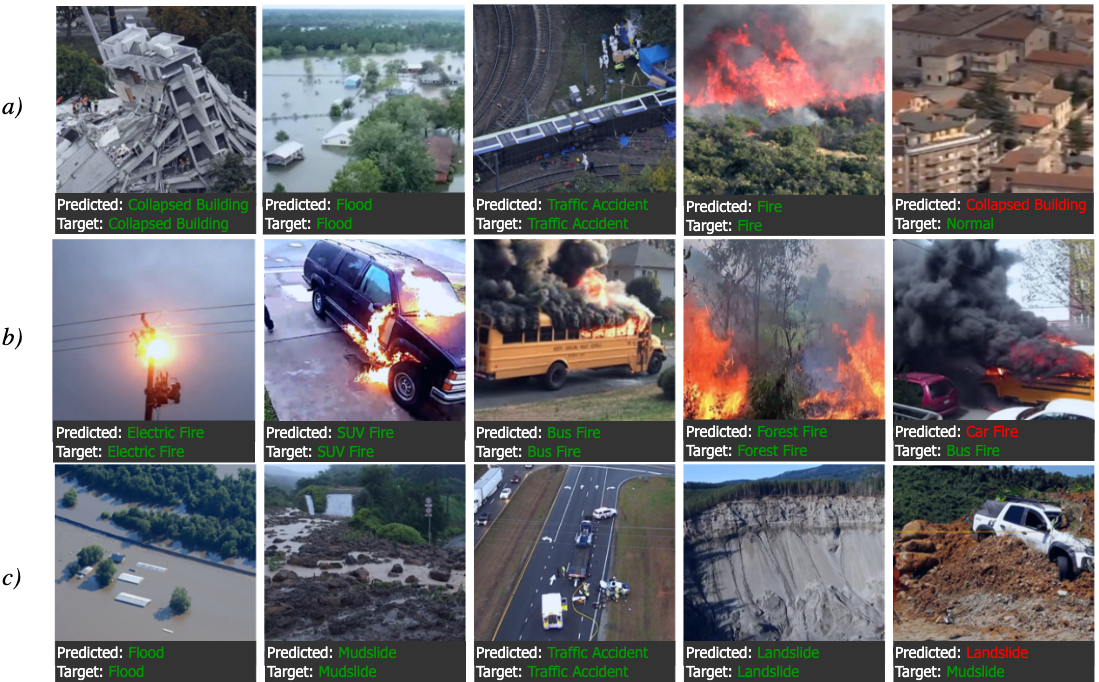} 
    \caption{Visual Results of the proposed model on benchmark datasets: a) AIDER, b) DFAN, and  c) DisasterEye.}
    \label{fig:inference}
\end{figure}

Usually, the model mistakes similar classes, such as landslide and mudslide, as seen from the last example in the third row. Moreover, the model makes errors because of its tendency to associate specific objects or features with particular classes. For example, the last picture in the first row shows scenery that can be associated with collapsed buildings. Finally, we observe failure cases when more than one disaster is illustrated in the photo. The last picture in the second row shows a bus fire, but cars are also present.
Increasing the number of frames for each category could potentially solve the problems mentioned above. Moreover, introducing multilabel classification per frame might reduce some of the issues.

\subsection{Ablation Experiments}
In this section, we evaluate the performance of our model while employing different optimization techniques. In addition to the proposed TensorRT optimization, we use MinMax, EMA, OMSE, Percentile, and FQ-ViT quantization techniques. The results can be found in Table \ref{tab:opt}.

MinMax, EMA, OMSE, and Percentile optimization techniques were designed mainly for CNN architectures. Therefore, they focus on Convolutional, Linear, and MatMul modules while lacking solutions for handling Softmax, LayerNorm, and GELU layers. This means many layers stay in a 32-bit floating point, resulting in the mix-precision network and unchanged model size; however, a significant speed-up, enough for real-time performance, can be seen for all these methods. 

FQ-ViT, specially tailored for transformer-based architectures, introduces powers-of-two scale quantization and log-int quantization for LayerNorm and Softmax, respectively. However, FQ-ViT does not focus on GELU layers, leading to partial performance. Due to dequantized floating-point parameters during inference, efficient low-precision arithmetic units are not fully utilized, and, therefore, we get unsatisfactory model acceleration.

\begin{table}[h]
\renewcommand{\arraystretch}{1.2} 
\centering
\caption{Optimization methods used on the proposed model.}
\begin{tabular}{l@{\hskip 3pt}l@{\hskip 3pt}c@{\hskip 3pt}c@{\hskip 3pt}c@{\hskip 3pt}c}

\hline
\multirow{2}{*}{\textbf{Dataset}} & \multirow{2}{*}{\textbf{Method (w/a/att)}} & \multirow{2}{*}{\textbf{Accuracy}} & \textbf{Latency} & \multirow{2}{*}{\textbf{FPS}} & \textbf{Model Size} \\
 &  &  & \textbf{[ms]} &  & \textbf{[MB]} \\ \hline
AIDER & Original 32/32/32    & 0.9825 & 48.87 & 20.46 & 107  \\
 & MinMax 8/8/8      & 0.9789 & 37.01 & 27.02 & 107  \\
 & EMA 8/8/8      & 0.9825 & 37.51 & 26.66 & 107  \\
 & OMSE 8/8/8      & 0.9825 & 37.62 & 26.58 & 107  \\
 & Percentile 8/8/8      & 0.9818 & 37.34 & 26.78 & 107  \\
 & FQ-VIT 8/8/4    & 0.9839 & 55.90 & 17.89 & 107  \\
& TensorRT INT8 & 0.9770 & 4.23 & 236.41 & \textbf{36.76}  \\ 
 & \textbf{TensorRT FP16} & \textbf{0.9799} & \textbf{2.97} & \textbf{336.92} & 58.89  \\
\hline
DFAN & Original 32/32/32    & 0.9309 & 63.36 & 15.78 & 107  \\
 & MinMax 8/8/8      & 0.9096 & 40.64 & 24.61 & 107  \\
 & EMA 8/8/8      & 0.9016 & 40.16 & 24.90 & 107  \\
 & OMSE 8/8/8      & 0.9149 & 40.71 & 24.56 & 107  \\
 & Percentile 8/8/8 & 0.9149 & 39.69 & 25.20 & 107  \\
 & FQ-VIT 8/8/4    & 0.9149 & 59.63 & 16.77 & 107  \\
 & TensorRT INT8 & 0.9096 & 4.16 & 240.17 & \textbf{34.91}  \\ 
 & \textbf{TensorRT FP16} & \textbf{0.9122} & \textbf{3.27} & \textbf{305.43} & 58.85  \\\hline
DisasterEye & Original 32/32/32    & 0.9016 & 79.54 & 12.57 & 107  \\
 & MinMax 8/8/8      & 0.8748 & 40.83 & 24.49 & 107  \\
 & EMA 8/8/8      & 0.8819 & 40.74 & 24.55 & 107  \\
 & OMSE 8/8/8      & 0.9070 & 41.98 & 23.82 & 107  \\
 & Percentile 8/8/8      & 0.8962 & 40.38 & 24.76 & 107  \\
 & FQ-VIT 8/8/4    & 0.8855 & 59.70 & 16.75 & 107  \\ 
 & TensorRT INT8 & 0.8945 & 4.33 & 230.73 & \textbf{33.47}  \\
 & \textbf{TensorRT FP16}      & \textbf{0.8962} & \textbf{3.01} & \textbf{332.26} & 58.74  \\\hline
\end{tabular}

\label{tab:opt}
\end{table}

\subsection{Comparison With the State-of-the-Art}

Table \ref{table:results_sota} shows the performance of the state-of-the-art models and our proposed method.
Our model outperforms almost all previously trained networks in terms of f1-score on the AIDER dataset. In \cite{Ijaz2023}, the authors used class weights due to unbalanced data when training their MobileNet. This helped the model make better predictions. Our results, on the other hand, rely purely on the transformers' capability to capture valuable information from frames. Regarding FPS, our proposed model surpasses all lighter models that contain fewer parameters and smaller model sizes. Finally, it demonstrates real-time performance on resource-constrained devices such as the Jetson Nano.

The limited network, such as MobileNet, performs poorly when trained on the DFAN dataset. At the same time, our model achieves the highest performance in terms of f1 score. It outperforms other solutions such as \cite{yar2022optimized}, \cite{khan2023enhancing}, \cite{yar2023fire_detection}, \cite{yar2024efficient} on high computational GPU devices regarding speed and achieves real-time performance on resource-constrained devices. 

These results show that our proposed framework's scalability and adaptability make it a valuable tool for deploying real-time disaster detection solutions on resource-limited UAV platforms.

\begin{table*}[htbp]
\centering
\caption{Summary of results of previous models trained on our datasets and our compressed, proposed model.}
\begin{tabular}{>{\centering\arraybackslash}m{1.2cm}>{
\arraybackslash}m{3.5cm}>
{\centering\arraybackslash}m{4.0cm}>{\centering\arraybackslash}m{1.5cm}>{\centering\arraybackslash}m{1.5cm}>{\centering\arraybackslash}m{2cm}>{\centering\arraybackslash}m{1.5cm}}
\hline
\textbf{Dataset} & \textbf{Models} & \textbf{System specification}  & \textbf{F1-Score} & \textbf{FPS}  & \textbf{Number of Parameters [M]} & \textbf{Model Size [MB]} \\ \hline
AIDER & EmergencyNet \cite{Kyrkou} & ARM Cortex-A57 & 0.957  & 25 & 0.09  & 0.78  \\
& TinyEmergencyNet \cite{mogaka2024tinyemergencynet} &  NVIDIA Quadro
RTX 5000 GPU & 0.940  & - & \textbf{0.04}   & \textbf{0.15} \\
& MFEMANet \cite{bhadra2023mfemanet} & - & 0.970 & - & - & - \\
& MobileNet \cite{Ijaz2023} & Jetson Nano & 0.980 & 4 & 3.2 & 37.00 \\
& MobileNet Compressed \cite{Ijaz2023} & Jetson Nano & - & 71 & 3.2 & 7.00 \\
& \textbf{Our Model FP16} & NVIDIA T4 GPU 16 GB GPU & \textbf{0.980} & \textbf{336.92} & 27.5 & 58.89 \\ 
& \textbf{Our Model INT8} & NVIDIA T4 GPU 16 GB GPU & 0.977 & 236.41 & 27.5 & 36.76 \\ 
& \textbf{Our Model INT8} & Jetson Nano & 0.977 & 45.22 & 27.5 & 34.00 \\\hline
DFAN & DFAN \cite{yar2022optimized} & NVIDIA GPU 2070 12
GB GPU  &  0.870 & 70.55  & 23.9 & 83.63 \\
& DFAN Compressed \cite{yar2022optimized} & NVIDIA GPU 2070 12
GB GPU  &  0.860 & 125.33  & 23.9 & 41.09 \\
& MobileNet \cite{yar2022optimized} & NVIDIA GPU 2070 12
GB GPU & 0.810 & - & 3.2 & 37.00 \\
& MAFire–Net \cite{khan2023enhancing} & NVIDIA GeForce RTX-3090 GPU & 0.875 & 78.31 & - & 74.43 \\
& ADFireNet \cite{yar2023fire_detection}  & NVIDIA GeForce RTX-3090 GPU & 0.900 & 72.50 & 7.2 & 38.00 \\
& MobileNetV3+MSAM \cite{yar2024efficient} & NVIDIA GeForce RTX-3090 GPU & 0.906 & 75.15 & \textbf{3.2} & \textbf{25.20} \\
& \textbf{Our Model FP16} & NVIDIA T4 GPU 16 GB GPU & \textbf{0.912} & \textbf{305.43} & 27.5 & 58.85 \\
& \textbf{Our Model INT8} &  NVIDIA T4 GPU 16 GB GPU & 0.910 & 240.17 & 27.5 & 34.91 \\ 
& \textbf{Our Model INT8} &  Jetson Nano & 0.910 & 45.04 & 27.5 & 34.00 \\\hline
\end{tabular}

\label{table:results_sota}
\end{table*}


\section{Conclusion}
\label{con}
This work confirms the effectiveness of optimized transformer-based architectures for real-time disaster classification on UAVs. By utilizing advanced quantization techniques, the model substantially reduces memory footprint and latency while maintaining accuracy across various disaster types. 

The results show that the precision calibration of FP16 and INT8 in TensorRT gives real-time performance with increased FPS compared to the original model and reduced model size by 50\% and 70\%, respectively. The optimized proposed model maintains high accuracy when datasets have sufficient samples per category, such as AIDER, while datasets with many classes but limited samples per category, such as DFAN, may experience some accuracy drop. Finally, TensorRT significantly enhances the model performance on edge devices such as Jetson Nano, achieving real-time performance.

The proposed framework demonstrates robust classification capabilities in diverse disaster scenarios and highlights the potential of deploying transformer models on UAVs for a prompt and autonomous disaster response. In addition, the performance of the proposed model on our DisasterEye dataset highlights its potential as a benchmark for the classification of real-life disaster scenarios. Future work will focus on extending the framework by adding more disaster scenarios and expanding the DisasterEye dataset.

\end{document}